\documentclass[twoside,leqno,twocolumn]{article}
\usepackage{ltexpprt}
\usepackage{graphicx}
\usepackage{amsfonts}
\usepackage{booktabs}
\usepackage{url}

\begin{document}

\title{\Large Learning Convolutional Text Representations for Visual Question Answering}
\author{Zhengyang Wang\thanks{School of Electrical Engineering and Computer
Science at Washington State University. Email: zwang6@eecs.wsu.edu}
\and Shuiwang Ji\thanks{School of Electrical Engineering and Computer
Science at Washington State University. Email: sji@eecs.wsu.edu}}
\date{}

\maketitle


\fancyfoot[R]{\footnotesize{\textbf{Copyright \textcopyright\ 2018 by SIAM\\
Unauthorized reproduction of this article is prohibited}}}





\begin{abstract} \small\baselineskip=9pt
Visual question answering (VQA) is a recently proposed artificial
intelligence task that requires a deep understanding of both images and
texts. In deep learning, images are typically modeled through convolutional
neural networks (CNNs) while texts are typically modeled through recurrent
neural networks (RNNs). In this work, we perform a detailed analysis on the
natural language questions in VQA, which raises a different need for text
representations as compared to other natural language processing tasks. Based
on the analysis, we propose to rely on CNNs for learning text
representations. By exploring various properties of CNNs specialized for text
data, we present our ``CNN Inception + Gate'' model for
text feature extraction in VQA. The experimental results show that simply
replacing RNNs with our CNN-based model improves question representations and thus the
overall accuracy of VQA models. In addition, our model has much fewer
parameters and the computation is much faster. We also prove that the text
representation requirement in VQA is more complicated
and comprehensive than that in conventional natural language processing
tasks. Shallow models like the fastText model, which can obtain comparable
results with deep learning models in simple tasks like text classification,
have poor performances in VQA.
\end{abstract}

\section*{Keywords}
Deep learning, visual question answering, convolutional neural networks, text representations

\section{Introduction.}

\begin{figure}[t]
\centering
\includegraphics[width=2.9in]{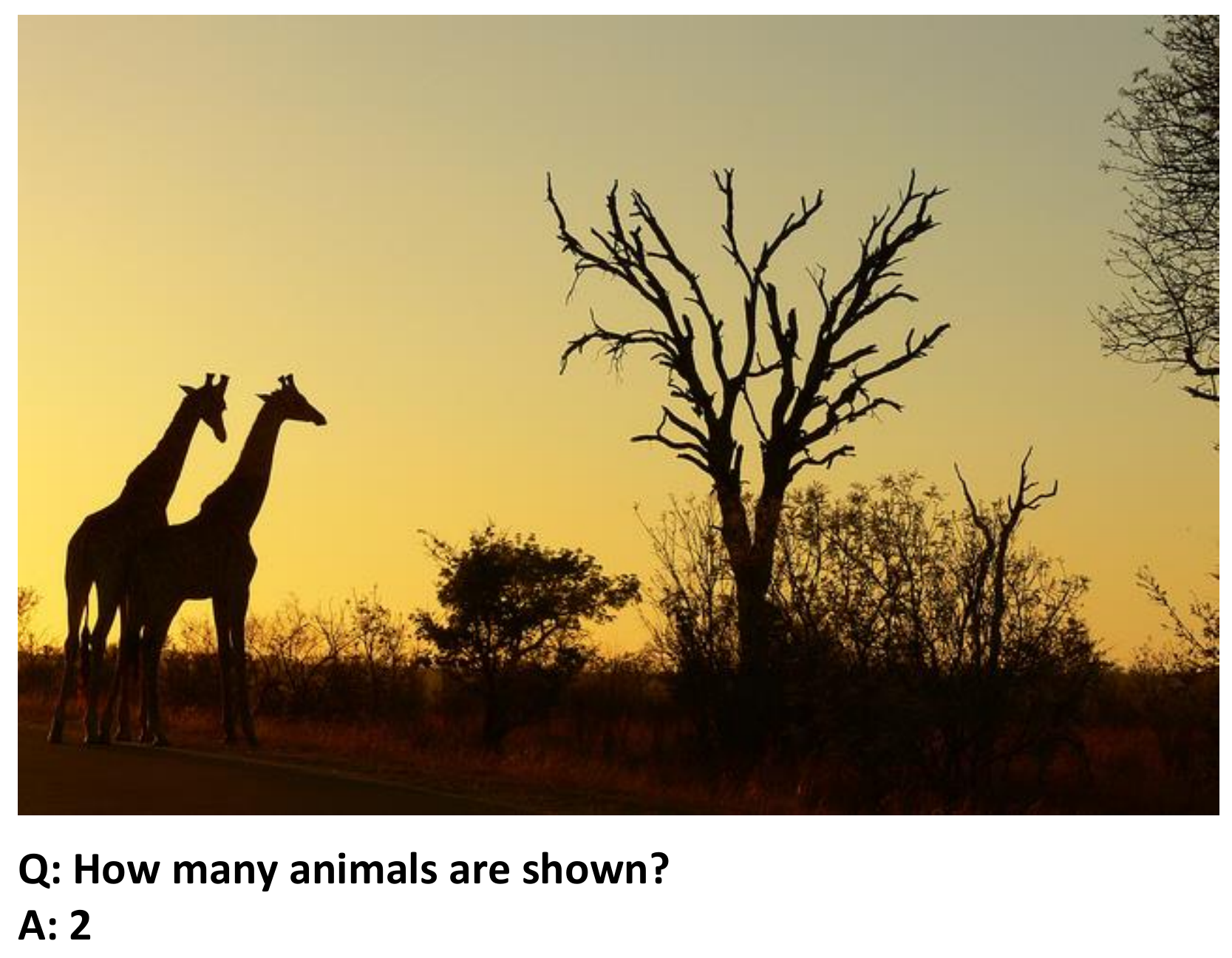}\vspace{-1pc}
\caption{An example of VQA task.}\label{fig:vqa}
\vspace{-1pc}
\end{figure}

Visual question answering (VQA)~\cite{antol2015vqa,kafle2016visual} asks an
agent to generate an accurate answer to a natural language question that
queries an image (Figure~\ref{fig:vqa}). This composite task involves a
variety of artificial intelligence fields, such as computer vision, natural
language processing, knowledge representation and reasoning. With the great
success of deep learning in these fields, an effective VQA agent can be built
with applications of artificial neural networks. A typical design is to use
an answer generator based on a joint representation of image and text
inputs~\cite{antol2015vqa}. A considerable body of research has been
conducted on how to efficiently combine image and text
representations~\cite{fukui2016multimodal,Kim2017,xiong2016dynamic,yang2015stacked,lu2016hierarchical},
while the fundamental question of learning these representations specifically
for VQA has not generated a lot of interests. In this work, we perform a
detailed analysis on text data in VQA and design text representation learning
methods that are appropriate for this task.

In VQA, the subtask of extracting visual information can be well addressed by
models commonly used in computer vision tasks like object
detection~\cite{lin2015bilinear,he2015deep} and image
classification~\cite{lecun1998gradient,he2015deep,szegedy2015going}, because
they share a similar requirement for image representations. Deep convolutional
neural networks (CNNs)~\cite{lecun1998gradient} have achieved significant
breakthroughs in computer vision and can be directly used in VQA. In natural
language processing, recurrent neural networks
(RNNs)~\cite{rumelhart1986sequential} are widely used to learn text
representations in tasks like sentiment
classification~\cite{kim2014convolutional,dos2014deep,zhang2015character,conneau2016very},
language modeling~\cite{dauphin2016language,mikolov2010recurrent}, and
machine translation~\cite{bahdanau2014neural,wu2016google,sutskever2014sequence}.
Parallel to image representations, most previous VQA
models~\cite{antol2015vqa,fukui2016multimodal,Kim2017,xiong2016dynamic,yang2015stacked,lu2016hierarchical}
directly rely on RNNs to extract textual information. However, our detailed
analysis reveals some special properties of text data in VQA, which indicates
that RNNs may not be the best fit for learning text representations in VQA.

With the above analysis and insight, we propose to apply CNNs for learning
text representations in VQA. Our experiments show that a very simple
CNN-based model outperforms a RNN-based model that has much more
parameters, a result consistent with our analysis. We further
explore techniques from CNNs for images and make specialized improvements
to build more effective models. Different methods for text vectorization
are also tested and analyzed. Our best model yields a substantial
improvement as compared to VQA models with RNN-based text models.

Our analysis also demonstrates a higher requirement for text representations
in VQA than that in traditional text tasks. Recent study on text
classification showed that a shallow model named fastText
\cite{joulin2016bag} can achieve comparable accuracies with deep learning models with
much faster computation. It is speculated that simple text classification
only needs shallow representation power. To validate our analysis, we conduct
experiments on learning text representations using fastText in VQA and
observe a significant decrease in accuracy. As a result, employing deep
models to learn text representations is more appropriate.

\section{Background.}

In this section, we describe convolutional neural networks, recurrent
neural networks and a common design pattern of VQA models.


\subsection{Convolutional Neural Networks and Recurrent Neural Networks.}\label{sec:cnn}

Convolutional neural networks (CNNs), which apply convolutional kernels in
artificial neural networks, have outperformed many other methods in computer
vision tasks. Unlike in image data processing, where convolutional kernels
are hardwired to be primitive feature detectors, CNNs train the parameters of
kernels, deciding what kinds of features are important to specific tasks. By
stacking several convolution layers, CNNs extract a hierarchy of increasingly
high-level image features. These features are then used as inputs to a
classifier, a text generator, or a decoder, depending on tasks. CNNs are
considered as a natural choice for matrix data like images which have fixed sizes.

Recurrent neural networks (RNNs) are developed for processing sequential
data. In natural language processing, text data is naturally a type of
sequential data so that RNNs are widely used in text tasks. However, recent
studies have shown that applying CNNs on sequential data with appropriate
pooling layers is feasible and able to obtain comparable results~\cite{kim2014convolutional,dos2014deep,zhang2015character,conneau2016very,dauphin2016language,johnson2014effective}.
In this work, we look into text data in VQA and propose a CNN-based model to
learn text representations based on our analysis and the characteristics of CNNs.
Details are discussed in Section~\ref{sec:main}.

\subsection{Visual Question Answering.}\label{sec:vqa}

Visual question answering (VQA) is considered as an advanced AI task, since
both visual and textual understanding and knowledge reasoning are needed in
VQA. Deep learning has shown its power in a variety of AI tasks. However,
training deep learning models demands a large amount of data. To this end,
various datasets aimed at VQA are collected and
published~\cite{kafle2016visual}. In~\cite{antol2015vqa}, a VQA dataset
(COCO-VQA) with a well-defined quantitative evaluation metric was made
available.

Most current VQA models share a similar design pattern. That is,
they consist of four basic components: an image feature extractor, a
text feature extractor, a feature combiner and a classifier. Image feature
extractors are usually pre-trained CNN-based models for image classification,
such as ResNet~\cite{he2015deep}, and GoogLeNet~\cite{szegedy2015going}.
Better classification models yield better results when used in VQA models.
However, this is not the case on text side, as discussed in
Section~\ref{sec:main}. Currently, most text feature extractors are RNNs like
LSTMs~\cite{hochreiter1997long}. To the best of our knowledge, only a very
simple CNN has been tried in \cite{yang2015stacked} and, without careful
analysis and design, it only achieved similar performances as RNNs. This work
provides a wide exploration of CNN-based text feature extractors based on
detailed analysis and obtains considerably better results. For feature
combiners, most efforts in VQA research have been devoted to improving them
to get a better joint representation derived from image and text
representations. \cite{fukui2016multimodal} won the $2016$ VQA challenge on
COCO-VQA by proposing the multimodal compact bilinear (MCB) pooling, which
was further improved in \cite{Kim2017}. In addition, the attention
mechanism has been proved to be effective as part of the combiner with its
ability to guide feature extractors to extract more related
information~\cite{yang2015stacked,lu2016hierarchical}. Contrast to these
works, we address the more fundamental question of how to learn better text
representations specifically for VQA. Finally, as proposed in
\cite{antol2015vqa}, we can cast the VQA problem into a classification
problem, where the joint representation is used as the input to a classifier.

\section{Text Representations in VQA.}\label{sec:main}


\subsection{Analysis of Texts in VQA.}\label{sec:motivation}

Natural language questions in VQA are different from other text data in
several aspects. First, people tend to ask short questions, according to
different VQA datasets~\cite{kafle2016visual}. For example, the longest
question in the training set of COCO-VQA contains only $22$ words, and the
average length is $6.2$. Most questions have $4$ to $10$ words. Second, the
required level for text understanding in VQA differs from that in
conventional natural language processing tasks. For instance, in sentiment
analysis~\cite{kim2014convolutional}, the model only needs to tell whether
the sentiment is positive or negative. So it will focus on emotional words
but pay little attention to other contents. In VQA, however, in order to
answer a question, a comprehensive understanding is required since a question
can ask anything. As a result, text feature extractors in VQA should be more
powerful and collects comprehensive information from raw texts. Third,
questions are different from declarative sentences in terms of syntax. And in
VQA, words in a question are highly related to the contents of its
corresponding image.

Based on these properties, we argue that, as compared to RNNs, CNNs are the
better choice for text feature extraction in VQA. By analyzing how human
beings process questions, we observe that there are two keys in question
understanding: one is understanding the question type and the other is
catching objects mentioned in the question and the relationships among them.
In many cases, the question type, which is usually determined by the first a
few words, directly describes what the answer looks like~\cite{antol2015vqa}.
Answers to questions starting with ``Is the'', ``Is there'', ``Do'' are
typically ``yes'' or ``no''. ``What number'' and ``How many'' questions must
have numbers as answers. Questions beginning with ``What color'', ``What
animal'', ``What sport'' and so on all explicitly indicate their answers'
categories. Meanwhile, objects and their relationships are usually nouns and
prepositional phrases, respectively. They provide guidance on locating
answer-related facts in the image, which is the fundamental
part of the attention mechanism in VQA models.

Now the task of text feature extraction becomes clear; that is, to obtain a
feature vector consisting of information about the question type and objects
being queried. To be more specific, the text representation is supposed to
extract what the starting words, nouns as well as prepositional phrases
represent. Considering words and phrases as features of text, a model
specializing on feature detection should be an appropriate choice. RNNs like LSTMs do not have explicit feature detection units. In
contrast to convolutional connections in CNNs, the connections within and
between units in RNNs are mostly fully-connected.

To summarize, CNNs are conceptually more appropriate as text feature
extractors in VQA, which is also validated by our experiments. Additional
advantages provided by CNNs are fewer parameters and easy parallelization,
which accelerate training and testing and reduce the risk of
over-fitting.

\subsection{Transforming Text Data.}\label{sec:trans}

A challenge of applying CNNs on text data is how to convert raw texts in a
format that CNNs can take, as they are originally designed for fixed-size
matrix data like images. To apply CNNs on texts directly, we need to
represent text data in the same way as how image data are represented. An
image is typically stored as a $3$-dimensional tensor, where the three
dimensions correspond to height, width, and number of channels, respectively.
Each pixel of the image is represented as a $n$-component vector
corresponding to $n$ channels.

Inspired by the bag-of-words model in natural language processing, a
vocabulary is first built. The vocabulary can be either word-based that
contains words appearing in the texts, or character-based, which is fixed for
a particular language. It is also reasonable for the vocabulary to include
punctuation as single words or characters. With the vocabulary, each sentence
can be transformed into an pseudo image whose height equals to $1$ and width
is defined based on the vocabulary. For word-level representations, the width
is number of words in a sentence; for character-level representations, we
count the number of characters. For the third dimension, similar to pixels in
an image, if we can convert each word as a vector, the length of the vector
is the number of channels. The problem is then reduced to
word vectorization, which is usually done by one-hot vectorization.

To make it concrete, we take the word-based vocabulary as an example, and the
character-based case can be easily generalized in Section~\ref{sec:char}.
Given a vocabulary $V$, each word can be represented as a one-hot vector;
namely a $|V|$-component vector with one $1$ at the position corresponding to
the index of the word in $V$ and $0$s for other entries, where $|V|$ is the
size of $V$. With one-hot vectorization, the number of channels becomes
$|V|$. As a result, a sentence with $L$ words is treated as a $1\times L$
pseudo image with $|V|$ channels, and it can be given into CNNs directly by
modifying the height of convolutional kernels into $1$ correspondingly.

\begin{figure*}[h!]
\centering
\includegraphics[width=0.88\textwidth]{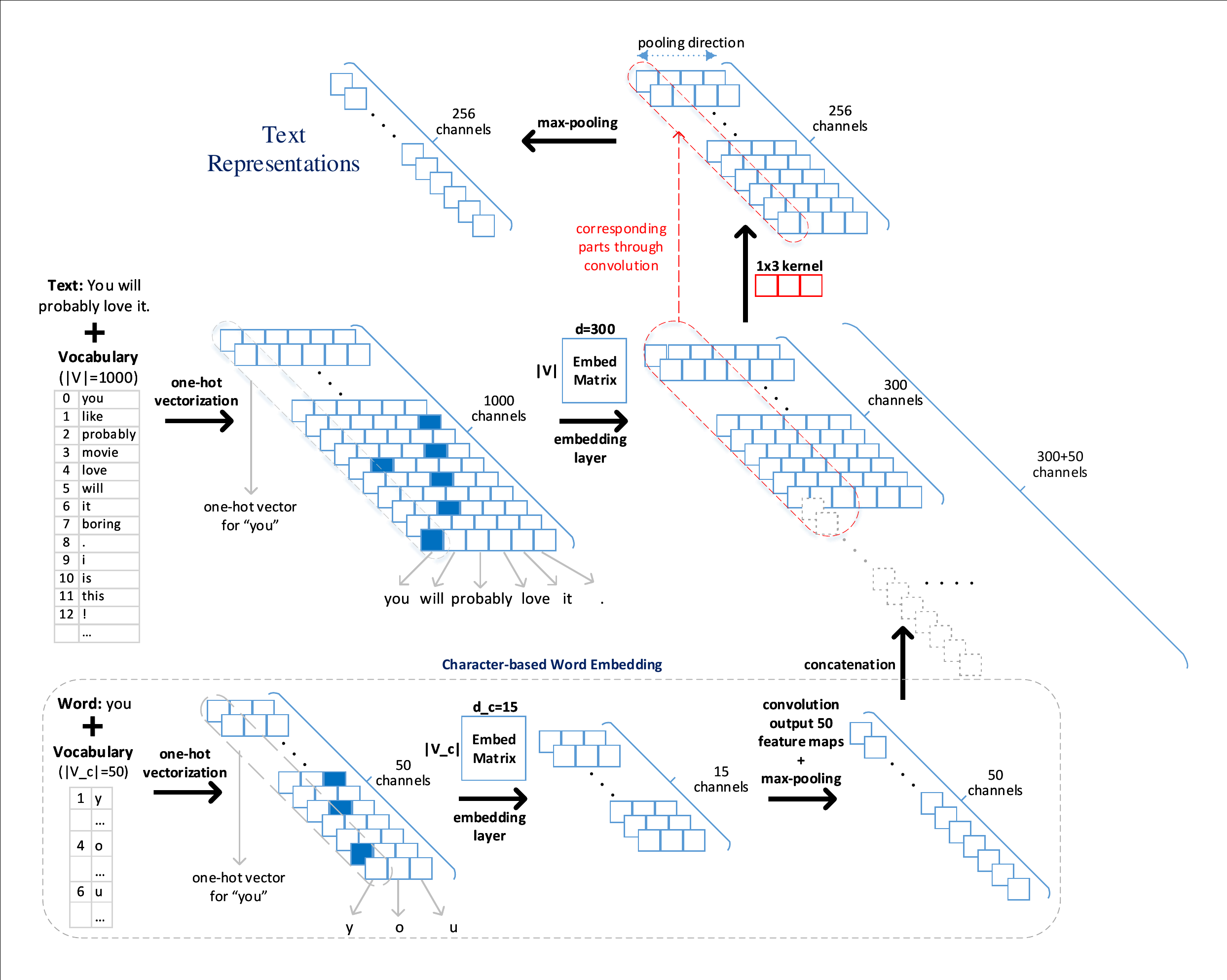}\vspace{-0.4cm}
\caption[LoF entry]{Illustration on employing CNNs to learn text
representations. Given a word-based vocabulary $V$, we first transform the
$L$-word sentence into a $1\times L$ image with $|V|$ channels by one-hot
vectorization. The blue units represent $1$ and white units represent $0$ for
this layer. Through the embedding layer, the number of channels is reduced to
$d$ (Section~\ref{sec:trans}). Then CNNs with $1$-D kernel can be directly
applied. After convolution, a max-pooling over the whole sentence is
performed to provide fix-sized inputs for the classifier
(Section~\ref{sec:pooling}). Wider and deeper convolution layers can be added
to this module easily.

The part in the dotted box illustrates the case where character-based
vocabulary is used to supplement word embedding vectors
(Section~\ref{sec:char}). In addition to the word-based vocabulary, a
character-based vocabulary $V\_c$ is provided. It transforms the
$C$-character word into a $1\times C$ image with $|V\_c|$ channels through
the same transformation. Then the embedding layer changes the number of
channels to $d\_c$. A CNN-based module followed by a global max-pooling
generates a word embedding, which is then concatenated to the word embedding
obtained from word-based vocabulary, generating the final pseudo image with
$d+d\_c$ channels. Note that here the CNN module is shared among different
words.}\label{fig:cnn}
\vspace{-1pc}
\end{figure*}

While one-hot embedding works well as inputs to CNNs in some
cases~\cite{johnson2014effective}, it is usually preferable to have a lower
dimensional embedding with two primary reasons. First, if $|V|$ is large,
which is usually the case for word-based vocabulary, computation efficiency
is low due to the sparsity and high dimensionality of inputs. Second, one-hot
embedding is semantically meaningless. Thus, an extra embedding layer is
usually inserted before CNNs. This layer maps the $|V|$-component vectors
into $d$-component vectors, where $d$ is much smaller than
$|V|$~\cite{kim2014convolutional,dos2014deep}. The embedding layer is
basically a multiplication of one-hot vectors with a $|V|\times d$ matrix to
perform a look-up operation. The embedding matrix can be trained as part of
the networks, which are task-specialized, or can be pre-trained using word
embedding like Word2Vec~\cite{mikolov2013distributed} or
GloVe~\cite{pennington2014glove}. Figure~\ref{fig:cnn} provides
a complete view of the transformations.

\subsection{Word-Based versus Character-Based Representations.}\label{sec:char}

Note that once a vocabulary is built, the remaining process to transform
texts follows the same path for different vocabularies. It is clear that the
vocabulary $V$ defines the pixels in the pseudo image. In the above example,
each word becomes a pixel. If the vocabulary is character-based, each
character, including space character and single punctuation, will be a pixel.

The main advantage of character-based vocabulary is that it produces much
longer inputs. This makes it possible for using deeper models. For long
texts, transforming text data using character-based vocabulary and applying
very deep CNNs leads to impressive
performances~\cite{conneau2016very,zhang2015character}. Another advantage is
that characters may include knowledge about how to form words. However, for
short texts, the size of the transformed data is still small even with
character-based vocabulary. Our experiments show that effective models for
long texts with character-based vocabulary fail to obtain high performances
in VQA (Section~\ref{sec:res}). It is believed that the inputs are too short
for the models to learn that space is the delimiter for words,
which is naturally given in word-based vocabulary case.

A combination of character-based and word-based vocabularies for short tests
has been explored in~\cite{dos2014deep} and achieved comparable results. In
this method, characters corresponding to each word are grouped together. Each
group of characters is transformed by character-based vocabulary and then fed
into a smaller model to generate a word vector. The word vector is then
concatenated with the corresponding word embedding from word-based vocabulary
to form a larger word representation. More details are given in
Figure~\ref{fig:cnn}. This method is also explored in our experiments.
Nevertheless, character-based vocabulary does not seem to be helpful.

\subsection{Handling Variable-Length Inputs.}\label{sec:pooling}

Another problem for text data is that each sentence is composed of different
numbers of words, which leads to variable-sized inputs and outputs of
convolution layers. However, the outputs of the whole CNN module are expected
to be fixed-sized, in order to serve as inputs to next module. Moreover, the
sizes of inputs to CNNs should also be consistent in consideration of
training.

Inspired by the pooling layers in CNNs for images~\cite{lecun1998gradient},
several pooling layers specialized for text data of variable lengths have
been proposed~\cite{kim2014convolutional,johnson2014effective}. We adopt the
method that applies one pooling for the whole sentence and selects the $k$
largest values instead of performing pooling locally. This is called $k$-max
pooling. By fixing $k$ for the last pooling layer of CNNs, the requirement
for fixed-sized outputs is satisfied. If $k=1$, it results in a global
max-pooling. More details are given in Figure~\ref{fig:cnn}.

While pooling layers can provide fixed-sized outputs regardless of the size
of inputs, fixed-sized inputs are also desired due to mini-batch training.
The solution is to perform padding and cropping. Cropping is usually used in
the case of long texts, especially with character-based vocabulary, which
simply cuts the part longer than a fixed length. For short texts like
questions in VQA, zero padding is typically used to pad each input to the
same length of the longest ones. This involves a problem that we only know
the longest length in the training set while there can be longer data. Thus
in practice, a combination of padding and cropping is used during testing.

\subsection{Deeper Networks for Short Texts.}\label{sec:res}

For long texts and images, deeper networks are important and beneficial.
Obstacles on going deeper are that very deep networks become hard to train
and suffer from the degradation problem. Residual networks
(ResNet)~\cite{he2015deep} overcame these obstacles by adding skip
connections from inputs to outputs of one layer or several layers. These skip
connections are named residual connections. They enable CNNs with hundreds of
layers to be trained efficiently and avoid the accuracy saturation problem.
Modified ResNet with $49$ layers for long texts has been explored in text
classification with character-based vocabulary~\cite{conneau2016very}.

We experiment with a ResNet with $8$ layers on texts in VQA with
character-based vocabulary. The results indicate that the inputs are too
short, and deeper networks suffer from over-fitting instead of training and
degradation problems. In fact, comparing to long texts where most samples
have more than $1000$ characters~\cite{conneau2016very} and multi-layer CNNs
work well, the length of texts in VQA is not enough for obtaining promising
outcomes from multi-layer CNNs. We also explore adding one residual block to
simple one-layer models but it also hurts the performances. It turns out
that, unlike mappings learned by intermediate layers in very deep models, the
mappings learned by the text feature extractor in VQA is not similar to
identity function, making the application of skip connections inappropriate.

These observations imply that CNNs on texts in VQA should not be deep. Our
experiments show that one-layer models achieved better performances.

\subsection{Wider Networks through Inception Modules.}\label{sec:incep}

Inception modules, proposed
by~\cite{szegedy2015going,kim2014convolutional}, involve combining
convolutional kernels of different sizes in one convolution layer.
This technique enables wider convolution layers. The motivation for
using inception modules for texts is straight-forward; that is,
different-sized kernels extract features from phrases of different
lengths. Based on this interpretation, the choice of the number of
kernels and their corresponding sizes should be data-dependent,
because different-sized phrases may have diverse importance in
various text data. We explore the settings and several improvements
in our experiments.

\subsection{Gated Convolutional Units.}\label{sec:gated}

LSTMs and GRUs improve RNNs by adding gates to control information
flow. In particular, the output gate controls information flow along
the sequential dimension. With this functionality, the output gate
can be used on any deep learning models.
In~\cite{van2016conditional} an output gate is also applied on CNNs.
Unlike LSTMs and GRUs that use fully-connected connections,
convolutional connections are used when generating output gates in
CNNs. Given an input to CNNs, which in our case is the transformed
data $I\in \mathbb{R}^{1\times L\times d}$ from text data, two
independent $1$-D convolutional kernels $K$ and $K_g$ are used to
form the output $O$ of the convolution layer as follows:
\begin{eqnarray}
g&=&\sigma(K_g\ast I+b_g),\label{eqn:1}\\
O&=&g\odot \tanh(K\ast I+b),\label{eqn:2}
\end{eqnarray}
where $g$ is the output gate, $\sigma$ is the \textit{sigmoid}
function, $\ast$ represents convolution, $\odot$ denotes
element-wise multiplication, $b$ and $b_g$ are bias terms. Gated
convolutional networks for language modeling was proposed
in~\cite{dauphin2016language}, and the activation function for the
original outputs was removed. That is, Eq.~(\ref{eqn:2}) is replaced
with
\begin{equation}\label{eqn:3}
O=g\odot (K\ast I+b).
\end{equation}
In our experiments, we explore both methods and combined gates with
inception modules, where different-sized kernels also generate
different gates. We achieve our best results with the method in
Eq.~(\ref{eqn:3}).

\subsection{fastText.}\label{sec:fast}

In~\cite{joulin2016bag} a shallow model named fastText was proposed, and it
achieved comparable results with deep learning models on several text
classification tasks with much less computation. In fastText, embedding
vectors of text data are directly averaged as sentence features. Formally, on
a word-based vocabulary, since the $1\times L$ pseudo image with $d$ channels
is actually a concatenation of $L$ $d$-component word vectors, the average
over $L$ word vectors results in a $d$-component sentence vector. This
sentence representation is given directly into the classifier. As compared to
deep learning models that use CNNs and RNNs, fastText obtains improvements in
terms of accuracy while achieving a $15,000$-fold speed-up due to the small
number of parameters.

The performance of fastText casts doubts on using deep learning models. However, it is
argued that simple text classification tasks may not take full advantage of
the higher representation power of deep learning~\cite{joulin2016bag}. As
stated in Section~\ref{sec:motivation}, the task of text understanding in VQA
is much more complicated and comprehensive. According to our experiments,
deep learning methods are superior to fastText in VQA, a result that is
consistent with our analysis.

\section{Experimental Studies.}

\subsection{General Settings.}\label{sec:experiment}

We report experimental results on COCO-VQA
dataset~\cite{antol2015vqa}\footnote{\url{http://visualqa.org/download.html}},
which consists of $204,721$ MSCOCO real images with $614,163$
questions. The data are divided into $3$ subsets: training ($82,783$
images with $248,349$ questions), validation ($40,504$ images with
$121,512$ questions) and testing ($81,434$ images with $244,302$
questions). In COCO-VQA, answers from ten different individuals are
collected for each question as ground truths. For training, the top
$K=3000$ frequent answers among all answers of the training set were
chosen to build the answer vocabulary. In each iteration, an
in-vocabulary answer is sampled as the label from ten ground truths
of each question. If all of the ten answers are out of the answer
vocabulary, the question is skipped. To evaluate the accuracy of a
generated answer, following evaluation metric was proposed
\cite{antol2015vqa}:
$$\mbox{Accuracy}=\min\left(\frac{\#\, \mbox{of humans with the answer}}{3}, 1\right),$$
where the generated answer is compared with each of the ten ground
truth answers, and the corresponding accuracy is computed. Since
evaluation on the testing set can only be processed on remote servers
during the VQA challenge~\cite{antol2015vqa}, and the testing labels
are not published, we choose to train and validate our models on the
training set only instead of the training+validation set
like~\cite{antol2015vqa,fukui2016multimodal}, and test on the
validation set.

Our baseline model is the challenge
winner~\cite{fukui2016multimodal}, which uses a $2$-layer LSTM as
the text feature extractor. This model is retrained on the training
set only. Meanwhile, unlike in~\cite{fukui2016multimodal}, we do not
use additional data sources like the pre-trained word embedding
(Word2Vec, GloVe) and other dataset (Visual
Genome~\cite{krishna2016visual}) to augment training. In order to
explore the power of models, we argue that additional data will
narrow the performance gap of different models. For comparison, we
only replace the LSTM text feature extractor with CNN models in
all experiments. All the results are reported in
Table~\ref{tab:result}. Our code is publicly
available\footnote{\url{https://github.com/divelab/vqa-text}}.

\begin{table}[t]
\centering
  \caption{Comparison of different text feature extractors. Accuracies per
    answer type are shown. Models are trained on the COCO-VQA training set and
    tested on the validation set. The retrained baseline model is shown as
    ``LSTM'' in Part $1$. The other parts are CNN-based
    models. ``Incep'', ``Res'', ``Bot'', ``G(A)'', ``G'', ``w'', ``c'' is short
    for ``Inception'', ``Residual'', ``Bottleneck'', ``Gate (tanh)'', ``Gate'',
    ``work'', ``char'', respectively.}
  \label{tab:result}
  \begin{tabular}{lcccc}
    \toprule
    \textbf{Models} & \textbf{Y/N} & \textbf{No.} & \textbf{Other} & \textbf{All}\\
    \midrule
    LSTM & 81.47 & 34.07 & 51.14 & 60.35 \\
    \midrule
    Non-Incep & 81.75 & 35.55 & 51.34 & 60.73 \\
    Incep (w) & 81.91 & \textbf{35.99} & 51.67 & 61.03 \\
    Incep + Res & 81.01 & 34.45 & 51.69 & 60.51 \\
    Incep + Bot & 80.12 & 35.51 & 50.58 & 59.74 \\
    Incep + G(A) & 82.09 & 35.47 & 51.84 & 61.10 \\
    Incep + G & \textbf{82.46} & 35.38 & \textbf{52.02} & \textbf{61.33} \\
    \midrule
    Incep (c) & 78.15 & 33.79 & 46.67 & 56.83 \\
    Deep Res & 77.19 & 33.39 & 46.09 & 56.14 \\
    \midrule
    Incep (c+w) & 82.05 & 35.39 & 51.43 & 60.88 \\
    \bottomrule
  \end{tabular}
\vspace{-1pc}
\end{table}

\subsection{Word-Based Models.}\label{sec:w}

Several CNN-based text feature extractors on word-based vocabulary
are implemented. The word-based vocabulary, which includes all words
that appear in the training set, has size $|V|=13321$. For word
embedding, we fix the dimension $d=300$. Dropout is applied on
text representations before they are given into next module. Part
$2$ in Table~\ref{tab:result} shows the results of these models.

``\textbf{Non-Inception}'' model is a one-layer model with one
$1\times 3$ convolutional kernel. With max-pooling over the whole
sentence, it produces a $2048$-component text vector
representation. This simple CNN-based model already outperforms the
baseline model, demonstrating that CNN-based model is better than
RNN-based one in VQA.

``\textbf{Inception (word)}'' model explores wider CNNs by replacing
the single $1\times 3$ kernel in ``CNN Non-Inception'' model with several
different-sized kernels in the same layer, as stated in
Section~\ref{sec:incep}. Different kernel settings are explored and
their results are given in Table~\ref{tab:incep}. Settings are named
in the format ``width of kernel (number of feature maps output by
this kernel)''. Note that the height of kernel is always $1$. The
resulting text vector representation has $2048$ components. All
these models outperform ``CNN Non-Inception'' model, showing that features
extracted from phrases of different lengths complement each other.
Table~\ref{tab:result} includes the best results. For all models
using inception modules, different kernel settings are explored. We
only report the best result for other models.

\begin{table}[t]
\centering
  \caption{Overall accuracies for ``CNN Inception (word)'' models with different
  kernel settings. Check Section~\ref{sec:w} for details.}
  \label{tab:incep}
  \begin{tabular}{lc}
    \toprule
    \textbf{Settings} & \textbf{Accuracy} \\
    \midrule
    2(512)+3(512)+4(512)+5(512) & \textbf{61.03} \\
    1(512)+3(512)+5(512)+7(512) & 60.96 \\
    3(1024)+5(512)+7(512) & 60.97 \\
    1(512)+3(1024)+5(512) & 60.95 \\
    3(1024)+5(1024) & 60.80 \\
  \bottomrule
\end{tabular}
\vspace{-1pc}
\end{table}

\begin{figure*}
\centering
\includegraphics[width=0.88\textwidth]{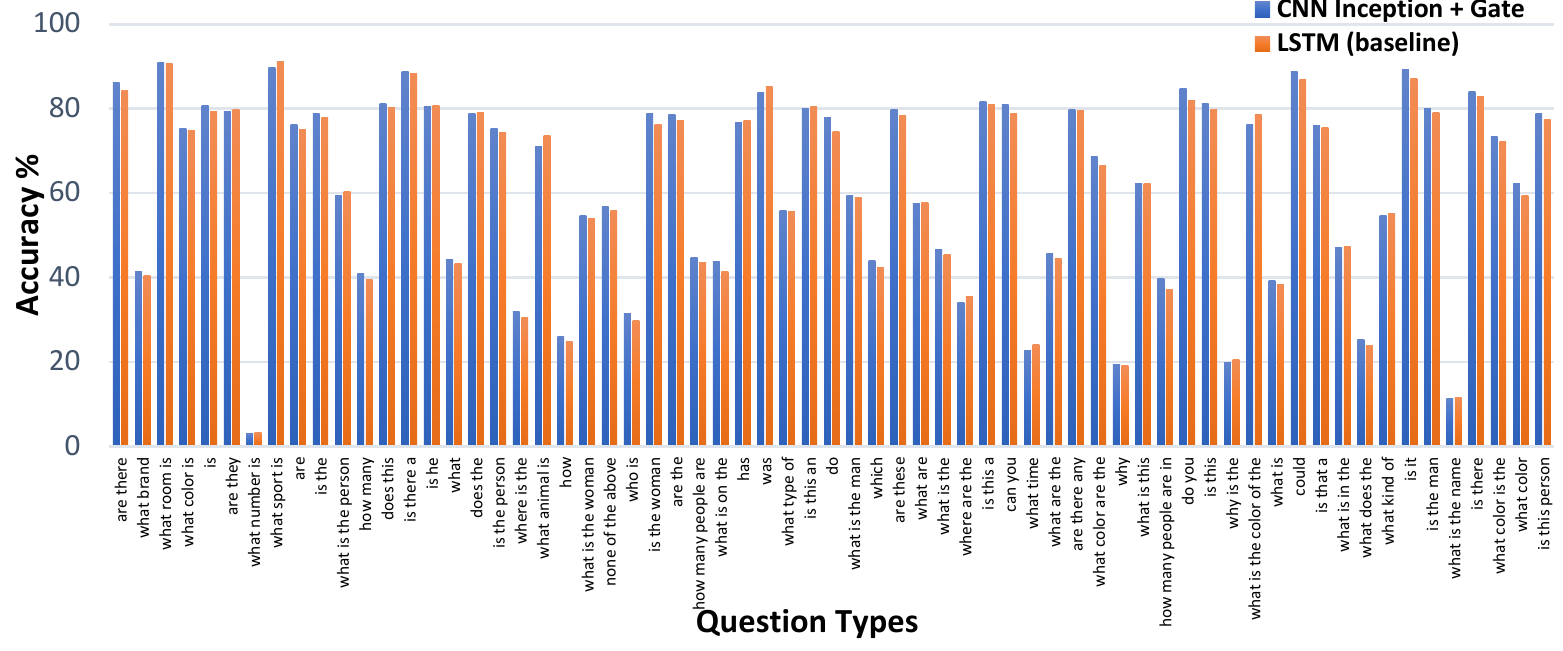}\vspace{-0.5cm}
\caption[LoF entry]{Comparison of accuracy per question type between
the ``Inception + Gate'' model and ``LSTM (baseline)''
model.}\label{fig:questype}
\vspace{-1pc}
\end{figure*}

``\textbf{CNN Inception + Residual}'' model tries going deeper. It
adds an identical layer with a residual connection from inputs to
outputs to ``CNN Inception (word)'' model (Section~\ref{sec:res}).
The best kernel setting is $1(512)+3(512)+5(512)+7(512)$. The extra
layer is supposed to further extract text features but hurt
performance in experiments. We conjecture that there is no need to
go deeper for the short inputs in VQA. Character-based vocabulary
will result in longer inputs and deeper models on it are discussed
in Section~\ref{sec:c}.

``\textbf{CNN Inception + Bottleneck}'' model is inspired by the
bottleneck architecture proposed by~\cite{he2015deep}. We apply
bottleneck on the convolution layer of ``CNN Inception (word)''
model with kernel setting $3(1024)+5(1024)$. For models on
image tasks, this architecture improves the accuracies while
reducing the number of parameters. However, it causes a significant
decrease in accuracy to our one-layer model for VQA, which indicates
that the bottleneck design is only suitable to very deep models.

``\textbf{CNN Inception + Gate (tanh)}'' model and
``\textbf{CNN Inception + Gate}'' model are CNN-based models with output
gates introduced in Section~\ref{sec:gated}, with Eqs.~(\ref{eqn:2})
and~(\ref{eqn:3}), respectively. Note that we combine the gate
architecture with the inception module: for each kernel $K$ in the
same convolution layer, there is a corresponding $K_g$. Both methods
improve ``CNN Inception (word)'' model by adding output gates. With
Eq.~(\ref{eqn:3}), we achieve our best text feature extractor with
$61.33\%$ accuracy. See Figure~\ref{fig:questype} for a comparison
in accuracy per question type between ``Inception + Gate'' model and
``LSTM (baseline)'' model. We can see for most question types,
``Inception + Gate'' model outperforms ``LSTM (baseline)'' model.

We compare the numbers of parameters of CNN-based text feature extractor with
LSTM-based ones in Table~\ref{tab:param}. CNN models improve the accuracy
with much fewer training parameters. This reduces the risk of over-fitting
and increases the speed.

\begin{table}[t]
\vspace{-1pc}
\centering
  \caption{The number of parameters for each model. We only compute the
  parameters of the text feature extractor.}
  \label{tab:param}
  \begin{tabular}{lc}
    \toprule
    \textbf{Models} & \textbf{Number of Parameters} \\
    \midrule
    LSTM (baseline) & 13,819,904 \\
    CNN Non-Inception & 1,845,248 \\
    CNN Inception (word) & 2,152,448 \\
    CNN Inception + Gate & 4,304,896 \\
  \bottomrule
\end{tabular}
\vspace{-1pc}
\end{table}

\subsection{Character-Based Models.}\label{sec:c}

Results for models that involve character-based vocabulary are
reported in parts $3$ and $4$ in Table~\ref{tab:result}. The two
models in part $3$ use character-based vocabulary only, while the
model in part $4$ uses a combination of both vocabularies
(Section~\ref{sec:char}). The character-based vocabulary collects
$|V\_c|=45$ characters: all lowercase characters in English,
punctuation as well as the space character. The kernel settings for
both inception-like models below are $2(512)+3(512)+4(512)+5(512)$.
Dropout is also applied.

``\textbf{CNN Inception (char)}'' model applies the same inception
module as ``CNN Inception (word)'' model but replaces the word-based
inputs with character-based inputs. The accuracy drops drastically.
As explained in Section~\ref{sec:char}, it is due to the short
length of the inputs, which is not enough for the model to learn how
to separate characters into words.

``\textbf{CNN Deep Residual}'' model attempts to take advantage of the longer
inputs provided by character-based vocabulary. We stack $5$ convolution
layers with residual connections and $3$ local pooling layers to build a deep
model. Contrast to the results of~\cite{zhang2015character,conneau2016very},
the model fails to work well. Again, comparison indicates the input length as
the cause of failure.

``\textbf{CNN Inception (char+word)}'' model makes use of both
word-based and character-based vocabularies as shown in
Figure~\ref{fig:cnn}. In our model, the characters of each word
generate a $150$-component word embedding vector, which is
concatenated with the $150$-component word embedding from word-based
vocabulary to form a $300$-component vector representing the word.
As compared to ``CNN Inception (word)'' model, it leads to a slight
accuracy decrease. This demonstrates that using character-based
vocabulary is not able to provide useful information from
constituent characters of the word. Based on these experiments, we
conclude that character-based vocabulary is not helpful in short
input cases like texts in VQA.

\subsection{Deep Learning Models versus fastText.}

As introduced in Section~\ref{sec:fast}, fastText is a shallow model that
achieves comparable results with deep learning models in
text classification tasks~\cite{joulin2016bag}. This result contradicts the
common belief that deep learning models can learn better representations. It has been
conjectured that the simple text classification task may not be the right one
to evaluate text representation methods. Given the higher requirements for
text understanding in VQA, we compare these models in VQA. In addition to the
original fastText model (``\textbf{fastText (word)}''), which averages word
embedding vectors to obtain sentence representations, we also explore
fastText (``\textbf{fastText (char+word)}'') with character-based vocabulary.
Similar to the idea in Section~\ref{sec:char}, character embedding of each
word is averaged to generate part of the word embedding. The results are
given in Table~\ref{tab:fast}. We can see the performance gap between deep
learning models and fastText. Clearly, it demonstrates the complexity of VQA
tasks and the power of deep learning.

\begin{table}[t]
\vspace{-1pc}
\centering
  \caption{Comparison of results between deep learning models and fastText.}
  \label{tab:fast}
  \begin{tabular}{lc}
    \toprule
    \textbf{Models} & \textbf{Accuracy} \\
    \midrule
    LSTM (baseline) & 60.35 \\
    CNN Inception + Gate & 61.33 \\
    fastText (word) & 59.30 \\
    fastText (char+word) & 59.24 \\
  \bottomrule
\end{tabular}
\vspace{-1pc}
\end{table}

\section{Conclusions.}

We perform detailed analysis on texts in VQA and propose to employ CNNs as
text feature extractors in current VQA models. By incorporating recent
research achievements in CNNs for images, our best model improves text
representations and the overall accuracy. By comparing deep learning models
with the fastText, we show that the requirement for text understanding in VQA
is more comprehensive and complicated than simple tasks. Based on our
research, we believe that our proposed methods can be extensively used for
learning text representations in other tasks on text data of similar properties.

\section*{Acknowledgements.}

This work was supported in part by National Science Foundation grant
IIS-1633359.

\bibliographystyle{siamplain}
\bibliography{vqa-reference}

\end{document}